\documentclass[11pt,a4paper]{article}
\usepackage[hyperref]{acl2021}
\usepackage{times}
\usepackage{latexsym}

\usepackage{microtype}
\usepackage{amsmath}
\usepackage{amsfonts}
\usepackage{algorithm}
\usepackage{arydshln}
\usepackage[noend]{algpseudocode}
\usepackage{bm}
\usepackage{bbm}
\usepackage{booktabs}
\usepackage{balance}
\usepackage{color}
\usepackage{CJKutf8}
\usepackage{graphicx} 
\usepackage{graphics}
\usepackage{multirow}
\usepackage{makecell}
\usepackage{paralist}
\usepackage{subfig}
\usepackage{textcomp}
\usepackage{threeparttable}
\usepackage[normalem]{ulem}

\usepackage{cleveref}
\crefformat{section}{\S#2#1#3} 

\aclfinalcopy 
\def\aclpaperid{2243} 

\title{Self-Training Sampling with Monolingual Data Uncertainty \\for Neural Machine Translation}

\author{
Wenxiang Jiao$^\dagger$\thanks{~~Work was mainly done when Wenxiang Jiao was interning at Tencent AI Lab.}  ~~Xing Wang$^\ddagger$  ~~Zhaopeng Tu$^\ddagger$ ~~Shuming Shi$^\ddagger$  ~~Michael R. Lyu$^\dagger$ ~~Irwin King$^\dagger$ \\
{$^\dagger$Department of Computer Science and Engineering} \\
{The Chinese University of Hong Kong, HKSAR, China} \\
{$^\ddagger$Tencent AI Lab} \\
$^\dagger${\tt \{wxjiao,lyu,king\}@cse.cuhk.edu.hk} \\
$^\ddagger${\tt \{brightxwang,zptu,shumingshi\}@tencent.com} \\}

\date{}

\begin{document}
\maketitle
\begin{abstract}
 Self-training has proven effective for improving NMT performance by augmenting model training with synthetic parallel data. The common practice is to construct synthetic data based on a randomly sampled subset of large-scale monolingual data, which we empirically show is sub-optimal. In this work, we propose to improve the sampling procedure by selecting the most informative monolingual 
 sentences to complement the parallel data. To this end, we compute the uncertainty of monolingual sentences using the bilingual dictionary extracted from the parallel data.
 Intuitively, monolingual sentences with lower uncertainty generally correspond to easy-to-translate patterns which may not provide additional gains. Accordingly, we design an uncertainty-based sampling strategy to efficiently exploit the monolingual data for self-training, in which monolingual sentences with higher uncertainty would be sampled with higher probability. 
 Experimental results on large-scale WMT English$\Rightarrow$German and English$\Rightarrow$Chinese datasets demonstrate the effectiveness of the proposed approach. Extensive analyses suggest that emphasizing the learning on uncertain monolingual sentences by our approach does improve the translation quality of high-uncertainty sentences and also benefits the prediction of low-frequency words at the target side.\footnote{The source code is available at \url{https://github.com/wxjiao/UncSamp}}
\end{abstract}

\section{Introduction}
\label{sec:introduction}
Leveraging large-scale unlabeled data has become an effective approach for improving the performance of natural language processing (NLP) models~\cite{devlin2019bert,brown2020language,jiao2020exploiting}. As for neural machine translation (NMT), compared to the parallel data, the monolingual data is available in large quantities for many languages. Several approaches on boosting the NMT performance with the monolingual data have been proposed, e.g., data augmentation~\cite{Sennrich:2016:BT, zhang2016exploiting}, semi-supervised training~\cite{cheng2016semi,zhang2018joint,cai2021neural}, pre-training~\cite{siddhant2020leveraging,liu2020multilingual}. Among them, data augmentation with the synthetic parallel data~\cite{Sennrich:2016:BT,Edunov:2018:understanding} is the most widely used approach due to its simple and effective implementation. It has been a de-facto standard in developing the large-scale NMT systems~\cite{hassan2018achieving,ng2019facebook,wu-EtAl:2020:WMT2,transmart2021}.

Self-training~\cite{zhang2016exploiting} is one of the most commonly used approaches for data augmentation. Generally, self-training is performed in three steps: (1) randomly sample a subset from the large-scale monolingual data; (2) use a ``teacher'' NMT model to translate the subset data into the target language to construct the synthetic parallel data; (3) combine the synthetic and authentic parallel data to train a ``student'' NMT model. 
Recent studies have shown that synthetic data manipulation~\cite{Edunov:2018:understanding,Caswell:2019:tagged} and training strategy optimization~\cite{Wu:2019:exploiting,Wang:2019:improving} in the last two steps can boost the self-training performance significantly. However, how to efficiently and effectively sample the subset from the large-scale monolingual data in the first step has not been well studied.


Intuitively, self-training simplifies the complexity of generated target sentences~\cite{kim2016sequence,Zhou:2019:understanding,Jiao:2020:data}, and easy patterns in monolingual sentences with deterministic translations may not provide additional gains over the self-training ``teacher'' model~\cite{shrivastava2016training}. Related work on computer vision also reveals that easy patterns in unlabeled data with the deterministic prediction may not provide additional gains~\cite{mukherjee2020uncertainty}. In this work, we investigate and identify the uncertain monolingual sentences which implicitly hold difficult patterns and exploit them to boost the self-training performance. Specifically, we measure the uncertainty of the monolingual sentences by using a bilingual dictionary extracted from the authentic parallel data (\cref{sec:identification_unc}). Experimental results show that NMT models benefit more from the monolingual sentences with higher uncertainty, except on those with excessively high uncertainty (\cref{sec:uncertain_part_performance}). By conducting the linguistic property analysis, we find that extremely uncertain sentences contain relatively poor translation outputs, which may hinder the training of NMT models (\cref{sec:uncertain_part_property}). 

Inspired by the above finding, we propose an uncertainty-based sampling strategy for self-training, in which monolingual sentences with higher uncertainty would be selected with higher probability
(\cref{sec:sampling_strategy}). Large-scale experiments on WMT English$\Rightarrow$German and English$\Rightarrow$Chinese datasets show that self-training with the proposed uncertainty-based sampling strategy significantly outperforms that with random sampling (\cref{sec:unconstrained}). 
Extensive analyses on the generated outputs confirm our claim by showing that our approach improves the translation of uncertain sentences and the prediction of low-frequency target words (\cref{sec:analysis}).

\paragraph{Contributions.} Our main contributions are:
\begin{itemize}
\item We demonstrate the necessity of distinguishing monolingual sentences for self-training.

\item We propose an uncertainty-based sampling strategy for self-training, which selects more complementary sentences for the authentic parallel data.

\item We show that NMT models benefit more from uncertain monolingual sentences in self-training, which improves the translation quality of uncertain sentences and the prediction accuracy of low-frequency words.

\end{itemize}

\section{Observing Monolingual Uncertainty}
\label{sec:observing}

In this section, we aimed to understand the effect of uncertain monolingual data on self-training. We first introduced the metric for identifying uncertain monolingual sentences, then the experimental setup and at last our preliminary results. 

\paragraph{Notations.} 
Let $X$ and $Y$ denote the source and target languages, and let $\mathcal{X}$ and $\mathcal{Y}$ represent the sentence domains of corresponding languages. Let $\mathcal{B}=\{({\bf x}^i, {\bf y}^i)\}_{i=1}^{N}$ denote the authentic parallel data, where ${\bf x}^i\in \mathcal{X}$, ${\bf y}^i\in \mathcal{Y}$ and $N$ is the number of sentence pairs. Let $\mathcal{M}_x = \{{\bf x}^j\}_{j=1}^{M_x}$ denote the collection of monolingual sentences in the source language, where ${\bf x}^j\in \mathcal{X}$ and $M_x$ is the size of the set. Our objective is to obtain a translation model $f:\mathcal{X}\mapsto\mathcal{Y}$, that can translate sentences from language $X$ to language $Y$.

\subsection{Identification of Uncertain Data}
\label{sec:identification_unc}
\paragraph{Data Complexity.}
According to \newcite{Zhou:2019:understanding}, the complexity of a parallel corpus can be measured by adding up the translation uncertainty of all source sentences. Formally, the translation uncertainty of a source sentence $\bf x$ with its translation candidates can be operationalized as conditional entropy:
\begin{align}
    \mathcal{H}({\bf Y}|{\bf X=x}) &= - \sum_{{\bf y}\in\mathcal{Y}}p({\bf y}|{\bf x})\log{p({\bf y}|{\bf x})} \\
    & \approx \sum_{t=1}^{T_x}\mathcal{H}(y|x=x_t),\label{eq:corpus_uncertainty}
\end{align}
where $T_x$ denotes the length of the source sentence, $x$ and $y$ represent a word in the source and target vocabularies, respectively. Generally, a high $\mathcal{H}({\bf Y}|{\bf X=x})$ denotes that a source sentence $\bf x$ would have more possible translation candidates. 

Equation~(\ref{eq:corpus_uncertainty}) estimates the translation uncertainty of a source sentence with all possible translation candidates in the parallel corpus. It can not be directly applied to the sentences in monolingual data due to the lack of corresponding translation candidates. One potential solution to the problem is utilizing a trained model to generate multiple translation candidates. However, generation may lead to bias estimation due to the generation diversity issue~\cite{li2016simple,shu2019generating}. More importantly, generation is extremely time-consuming for large-scale monolingual data.

\paragraph{Monolingual Uncertainty.}
To address the problem, we modified Equation~(\ref{eq:corpus_uncertainty}) to reflect the uncertainty of monolingual sentences. We estimate the target word distribution conditioned on each source word based on the authentic parallel corpus, and then use the distribution to measure the translation uncertainty of the monolingual example. Specifically, we measure the uncertainty of monolingual sentences based on the bilingual dictionary.

For a given monolingual sentence ${\bf x}^{j}\in \mathcal{M}_x$, its uncertainty $\mathrm{U}$ is calculated as: 
\begin{align}
    \mathrm{U}({\bf x}^{j}|\mathcal{A}_b) = \frac{1}{T_x}\sum_{t=1}^{T_x}\mathcal{H}(y|\mathcal{A}_{b},x=x_t), \label{eq:monolingual_uncertainty}
\end{align}
which is normalized by ${T_x}$ to avoid the length bias. A higher value of $\mathrm{U}$ indicates a higher translation uncertainty of the monolingual sentence.

In Equation~\ref{eq:monolingual_uncertainty}, the word level entropy $\mathcal{H}(y|\mathcal{A}_{b},x=x_t)$ captures the translation modalities of each source word by using the bilingual dictionary $\mathcal{A}_b$. The bilingual dictionary records all the possible target words for each source word, as well as translation probabilities. It can be built from the word alignments by external alignment toolkits on the authentic parallel corpus. For example, given a source word $x$ with all three word translations $y_1$, $y_2$ and $y_3$ and the translation probabilities of $p(y_1|x)$, $p(y_2|x)$ and $p(y_3|x)$, respectively, the word level entropy can be calculated as follows:
\begin{align}
    \mathcal{H}(y|\mathcal{A}_b,x_i) = - \sum_{y_j\in\mathcal{A}_{b}(x_i)}p(y_j|x_i)\log{p(y_j|x_i)}.
\end{align}

\begin{figure}[t!]
    \centering
    \includegraphics[height=0.33\textwidth]{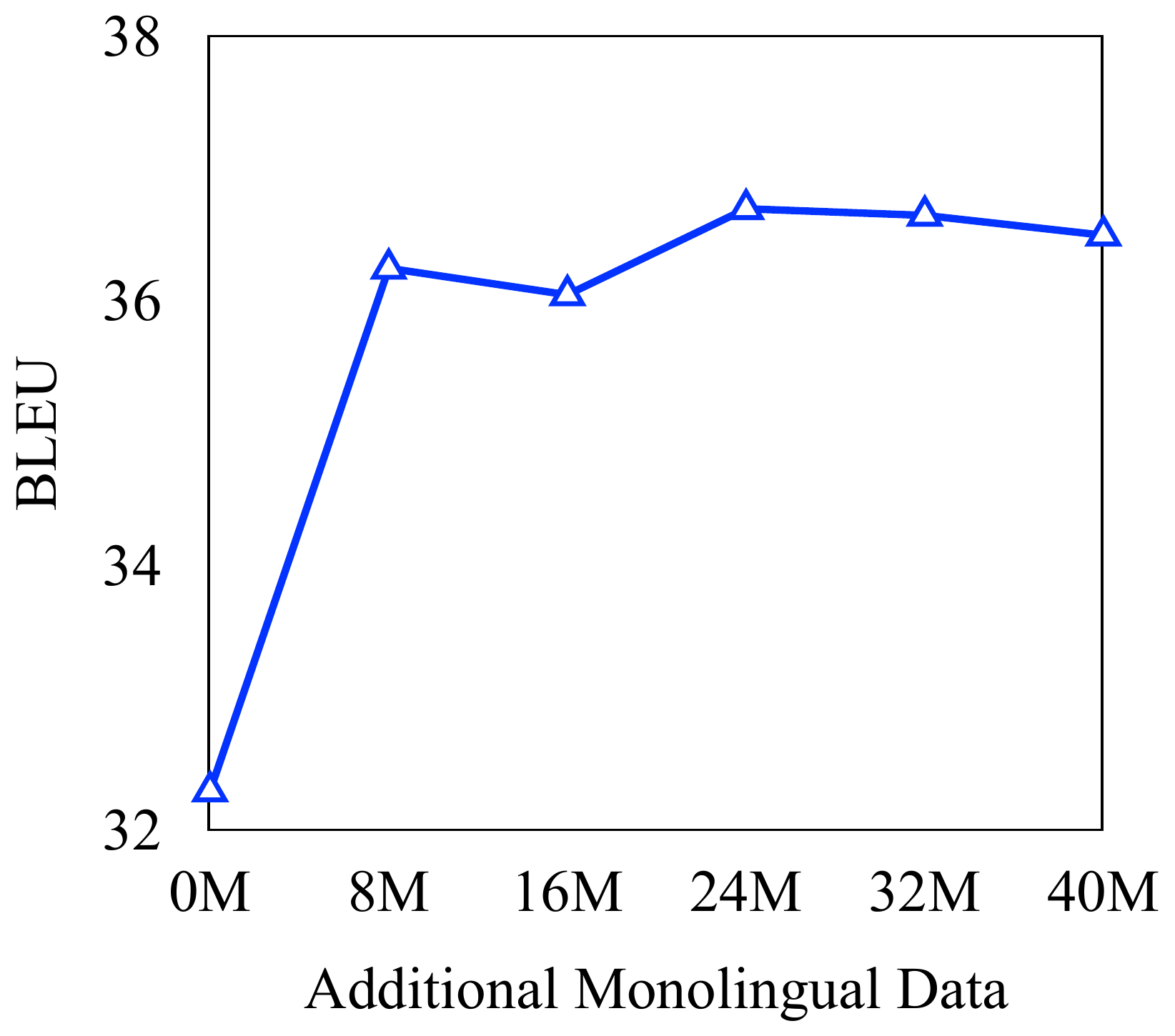}
    \caption{Performance of self-training with increased size of monolingual data. The BLEU score is averaged on WMT En$\Rightarrow$De newstest2019 and newstest2020.}
    \label{fig:self_training_bleu}
\end{figure}

\subsection{Experimental Setup}
\label{sec:setup}

\paragraph{Data.} 
We conducted experiments on two large-scale benchmark translation datasets, i.e., WMT English$\Rightarrow$German (En$\Rightarrow$De) and WMT English$\Rightarrow$Chinese (En$\Rightarrow$Zh). The authentic parallel data for the two tasks consists of about $36.8$M and $22.1$M sentence pairs, respectively. The monolingual data we used is from newscrawl released by WMT2020. We combined the newscrawl data from year 2011 to 2019 for the English monolingual corpus, consisting of about 200M sentences. We randomly sampled 40M monolingual data for En$\Rightarrow$De and 20M for En$\Rightarrow$Zh unless otherwise stated. 
We adopted newstest2018 as the validation set and used newstest2019/2020 as the test sets.
For each language pair, we applied Byte Pair Encoding~\cite[BPE,][]{Sennrich:2016:BPE} with 32K merge operations.


\paragraph{Model.} 
We chose the state-of-the-art \textsc{Transformer}~\cite{Vaswani:2017:NIPS} network as our model, which consists of an encoder of 6 layers and a decoder of 6 layers. 
We adopted the open-source toolkit Fairseq~\cite{Ott:2019:naacl} to implement the model. We used the \textsc{Transformer-Base} model 
for preliminary experiments (\cref{sec:uncertain_part_performance})
and the constrained scenario (\cref{sec:constrained}) for efficiency. For the unconstrained scenario (\cref{sec:unconstrained}), we adopted the \textsc{Transformer-Big} model. Results on these models with different capacities can also reflect the robustness of our approach.
For the \textsc{Transformer-Base} model, we trained it for 150K steps with 32K ($4096\times8$) tokens per batch.
For the \textsc{Transformer-Big} model, we trained it for 30K steps with 460K ($3600\times128$) tokens per batch with the cosine learning rate schedule~\cite{Wu:2019:pay}. We used 16 Nvidia V100 GPUs to conduct the experiments and selected the final model by the best perplexity on the validation set.

\paragraph{Evaluation.}
We evaluated the models by BLEU score~\cite{Papineni:2002:bleu} computed by SacreBLEU~\cite{Post:2018:sacrebleu}\footnote{
$\texttt{BLEU+case.mixed+lang.[Task]+numrefs.1}$ $\texttt{+smooth.exp++test.wmt[Year]+tok.[Tok]+ver}$ $\texttt{sion.1.4.14}$, Task=en-de/en-zh, Year=19/20, Tok=13a/zh}. For the En$\Rightarrow$Zh task, we added the option \texttt{--tok zh} to SacreBLEU. We measured the statistical significance of improvement with paired bootstrap resampling~\cite{Koehn:2004:emnlp} using \texttt{compare-mt}\footnote{\url{https://github.com/neulab/compare-mt}}~\cite{neubig2019:naacl}.

\begin{figure}[t!]
    \centering
    \includegraphics[height=0.34\textwidth]{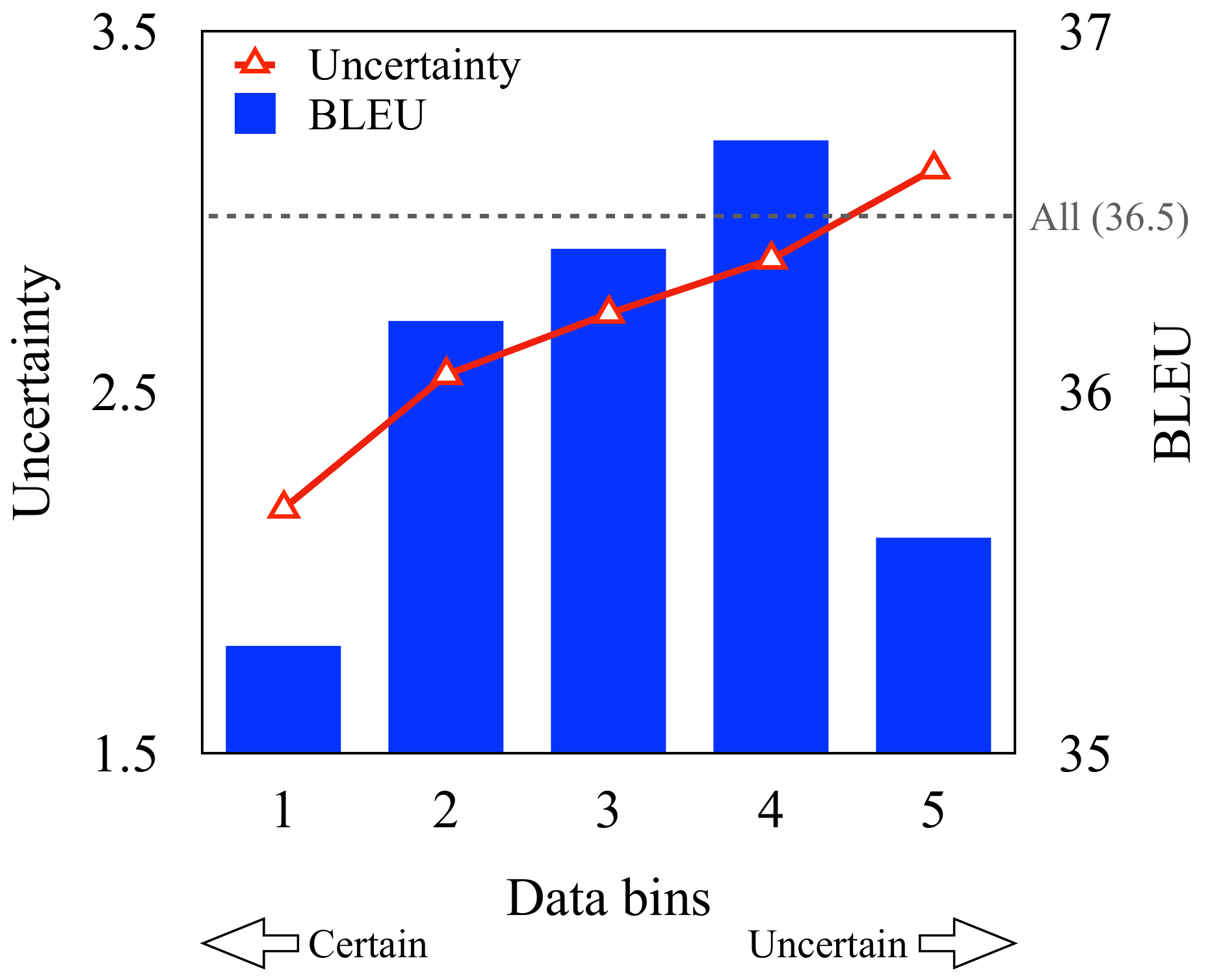}
    \caption{Relationship between uncertainty of monolingual data and the corresponding NMT performance. The BLEU score is averaged on WMT En$\Rightarrow$De newstest2019 and newstest2020.}
    \label{fig:uncertainty_bleu}
\end{figure}

\begin{figure*}[t!]
    \centering  
    \subfloat[Sentence Length]{
    \includegraphics[height=0.3\textwidth]{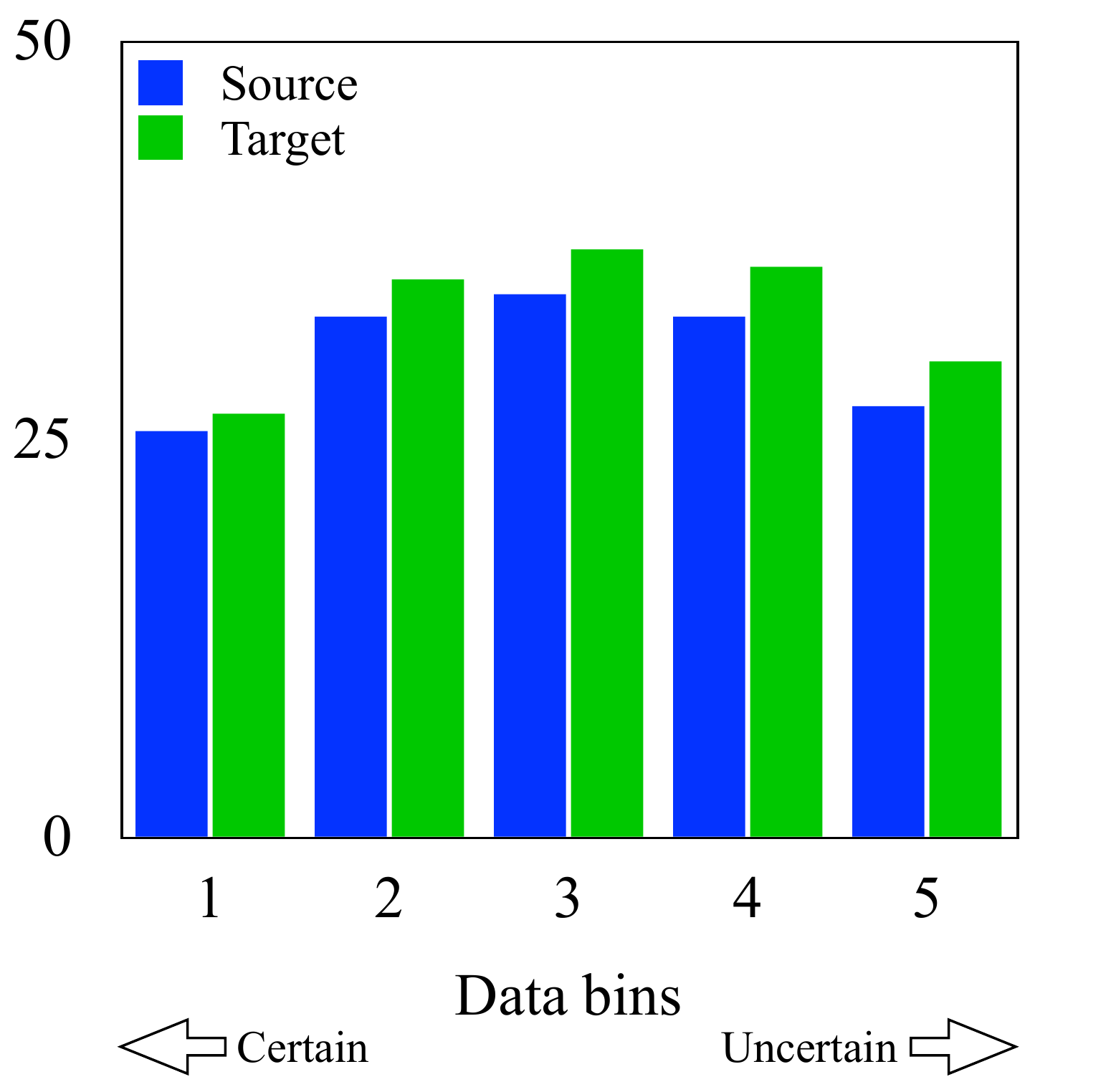}}
    \hspace{0.02\textwidth}
    \subfloat[Word Rarity]{
    \includegraphics[height=0.3\textwidth]{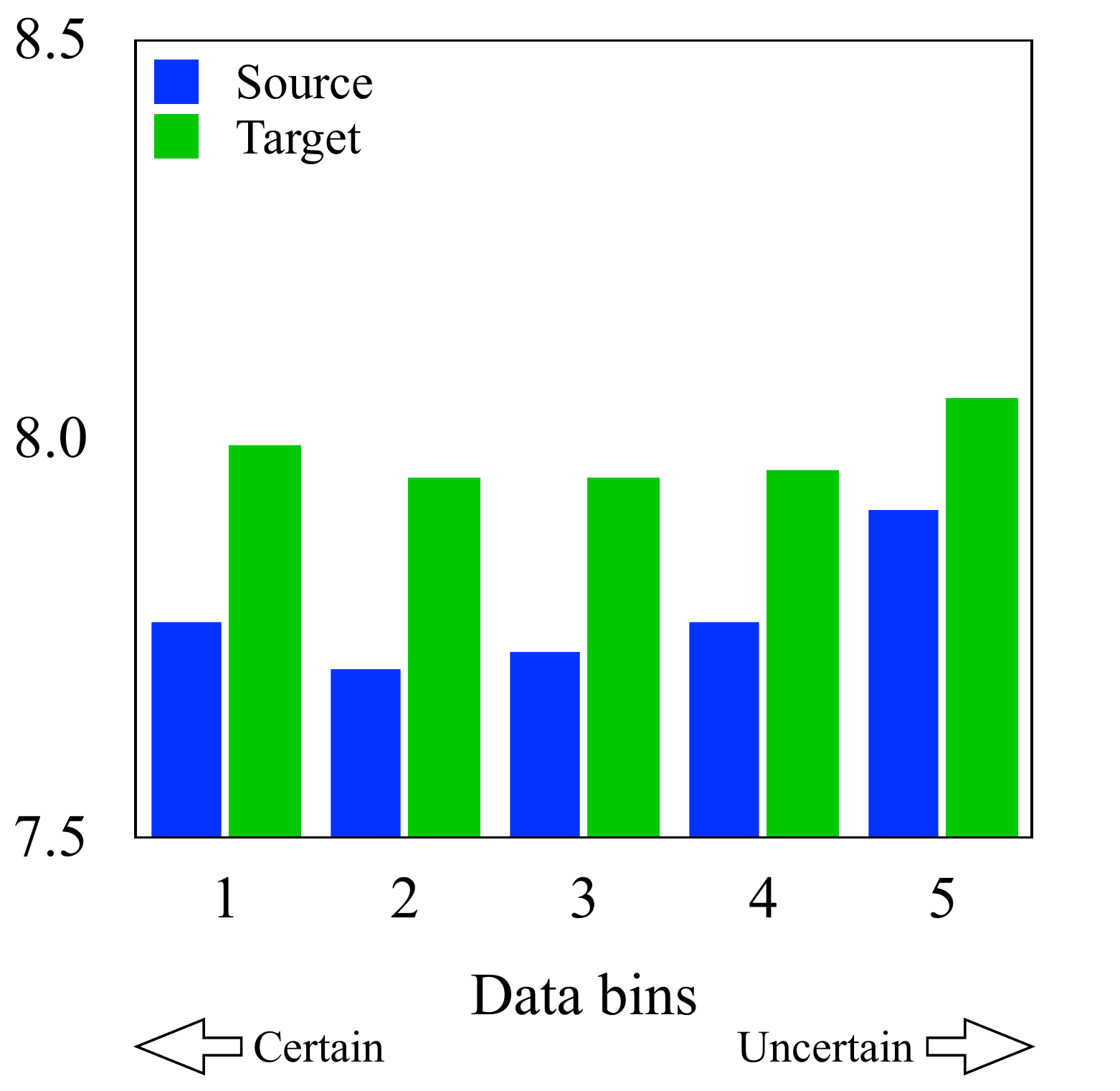}}
    \hspace{0.02\textwidth}
    \subfloat[Coverage]{
    \includegraphics[height=0.3\textwidth]{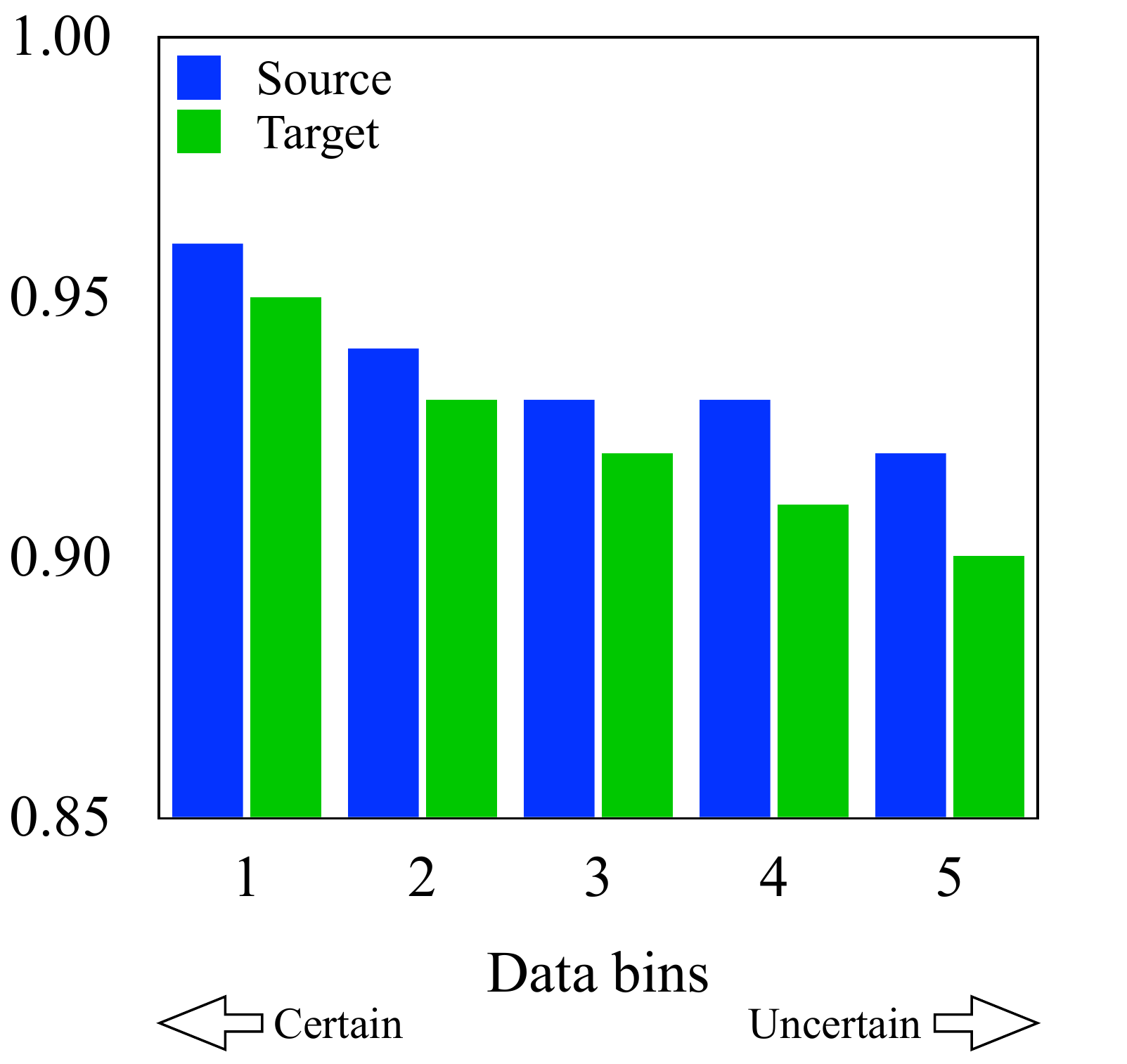}}
    \caption{Comparison of monolingual sentences with varied uncertainty in terms of three properties, including sentence length, word rarity, and coverage.}
    \label{fig:curves-property}
\end{figure*}

\subsection{Effect of Uncertain Data}
\label{sec:uncertain_part_performance}
First of all, we investigated the effect of monolingual data uncertainty on the self-training performance in NMT. We conducted the preliminary experiments on the WMT En$\Rightarrow$De dataset with the \textsc{Transformer-Base} model. 
We sampled 8M bilingual sentence pairs from the authentic parallel data and randomly sampled 40M monolingual sentences for the self-training.
To ensure the quality of synthetic parallel data, we trained a \textsc{Transformer-Big} model for translating the source monolingual data to the target language.  We generated translations using beam search with beam width $5$, and followed~\newcite{Edunov:2018:understanding}\footnote{\url{https://github.com/pytorch/fairseq/tree/master/examples/backtranslation}} to filter the generated sentence pairs (See Appendix~\ref{sec:appendix_syn}).

\paragraph{Self-training v.s. Data Size.}
We took a look at the performance of standard self-training and its relationship with data size.
Figure~\ref{fig:self_training_bleu} showed the results. Obviously, self-training with 8M synthetic data can already improve the NMT performance by a significant margin (36.2 averaged BLEU score on WMT En$\Rightarrow$De newstest2019 and newstest2020). Increasing the size of added monolingual data does not bring much more benefit. With all the 40M monolingual sentences, the final performance achieves only 36.5 BLEU points. It indicates that adding more monolingual data only is not a promising way to improve self-training, and more sophisticated approaches for exploiting the monolingual data are desired.


\paragraph{Self-training v.s. Uncertainty.}
In this experiment, we first adopted \textit{fast-align}\footnote{\url{https://github.com/clab/fast_align}} to establish word alignments between source and target words in the authentic parallel corpus and used the alignments to build the bilingual dictionary $\mathcal{A}_b$. Then we used the bilingual dictionary to compute the data uncertainty expressed in Equation~(\ref{eq:monolingual_uncertainty}) for the sentences in the monolingual data set. After that, we ranked all the 40M monolingual sentences and grouped them into 5 equally-sized bins (i.e., 8M sentences per bin) according to their uncertainty scores. At last, we performed self-training with each bin of monolingual data.

We reported the translation performance in Figure~\ref{fig:uncertainty_bleu}. 
As seen, there is a trend of performance improvement with the increase of monolingual data uncertainty (e.g., bins 1 to 4) until the last bin. The last bin consists of sentences with excessively high uncertainty, which may contain erroneous synthetic target sentences. Training on these sentences forces the models to over-fit on these incorrect synthetic data, resulting in the confirmation bias issue~\cite{arazo2020pseudo}. These results corroborate with prior studies~\cite{chang2017active,mukherjee2020uncertainty} such that learning on certain examples brings little gain while on the excessively uncertain examples may also hurt the model training.

\subsection{Linguistic Properties of Uncertain Data}
\label{sec:uncertain_part_property}

We further analyzed the differences between the monolingual sentences with varied uncertainty to gain a deeper understanding of the uncertain data. Specifically, we performed linguistic analysis on the five data bins in terms of three properties: 1) sentence length that counts the tokens in the sentence, 2) word rarity~\cite{Platanios:2019:competence} that measures the frequency of words in a sentence with a higher value indicating a more rare sentence, and 3) translation coverage~\cite{khadivi2005automatic} that measures the ratio of source words being aligned with any target words. The first two reflect the properties of monolingual sentences while the last one reflects the quality of synthetic sentence pairs. We also presented the results of the synthetic target sentences for reference. Details of the linguistic properties are in Appendix~\ref{sec:appendix_ling}.

The results are reported in Figure~\ref{fig:curves-property}. For the length property, we find that monolingual sentences with higher uncertainty are usually longer except for those with excessively high uncertainty (e.g., bin 5).  The monolingual sentences in the last data bin noticeably contain more rare words than other bins in Figure~\ref{fig:curves-property}(b), and the rare words in the sentences pose a great challenge in the NMT training process~\cite{gu2020token}. 
In Figure~\ref{fig:curves-property}(c), the overall coverage in bin 5 is the lowest among the self-training bins. In contrast, bin 1 with the lowest uncertainty has the highest coverage. 
These observations suggest that monolingual sentences in bin 1 indeed contain the easiest patterns while monolingual sentences in bin 5 are the most difficult ones, which may explain their relatively weak performance in Figure~\ref{fig:uncertainty_bleu}.

\section{Exploiting Monolingual Uncertainty}
\label{sec:exploting}
By analyzing the effect of monolingual data uncertainty on self-training in Section~\ref{sec:observing}, we understood that monolingual sentences with relatively high uncertainty are more informative while also with high quality, which motivates us to emphasize the training on these sentences. 
In this section, we introduced the uncertainty-based sampling strategy for self-training and the overall framework. 

\begin{figure}[t!]
    \centering  
    \subfloat[Uncertainty]{
    \includegraphics[height=0.26\textwidth]{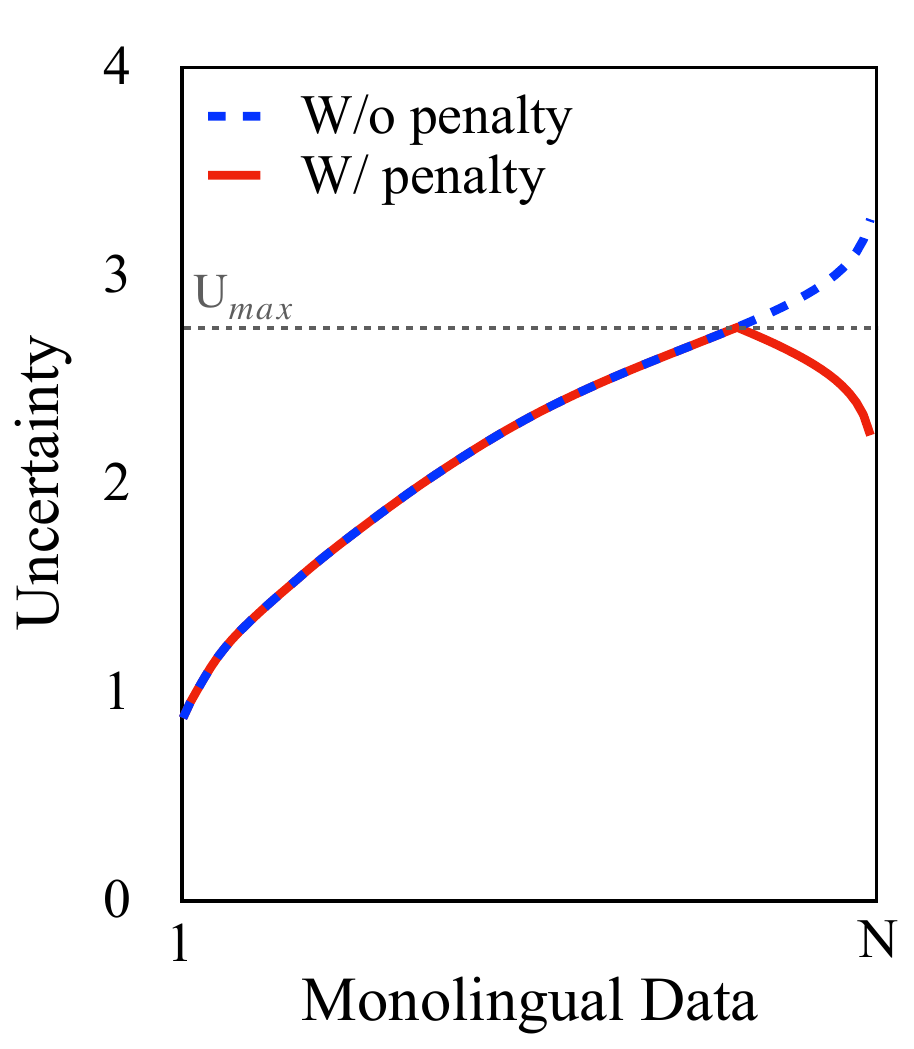}}
    \hspace{0.01\textwidth}
    \subfloat[Sampling Probability]{
    \includegraphics[height=0.26\textwidth]{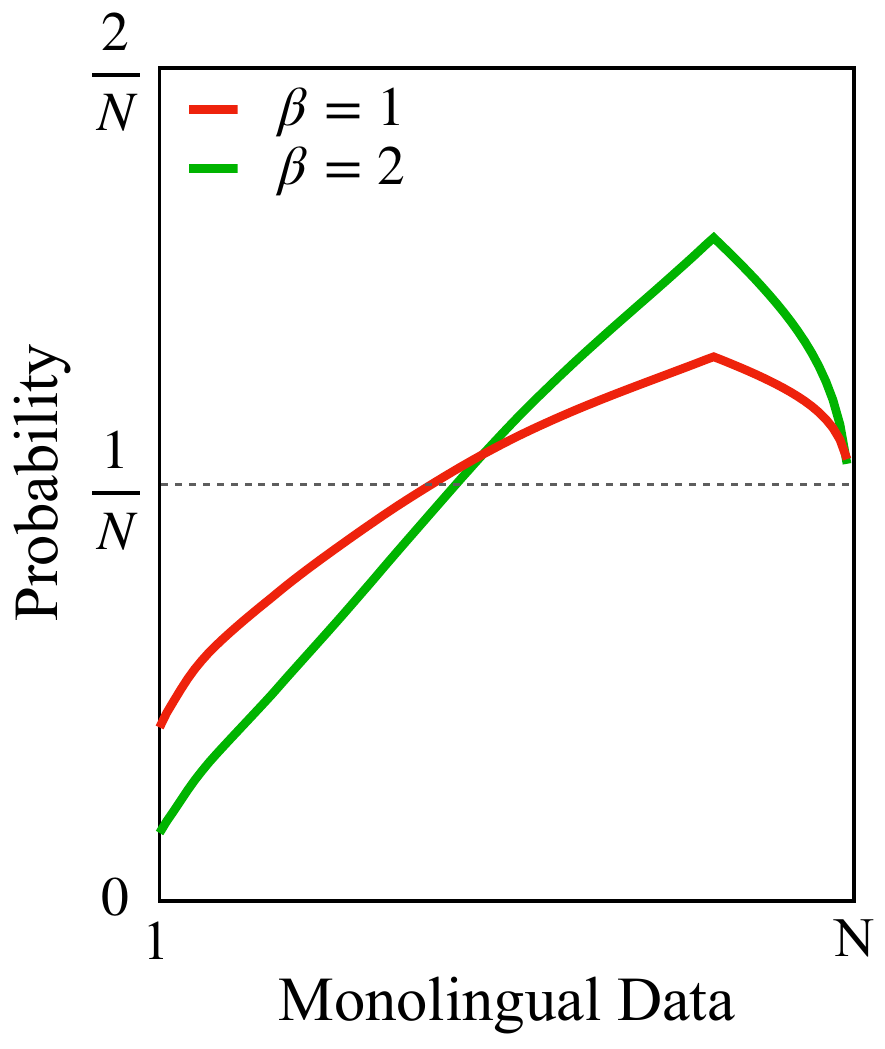}}
    \caption{Distribution of modified monolingual uncertainty and  sampling probability. The sample with high uncertainty has more chance to be selected while that with excessively high uncertainty would be penalized. }
    \label{fig:curves-sampprob}
\end{figure}

\begin{figure*}[t!]
    \centering
    \includegraphics[height=0.4\textwidth]{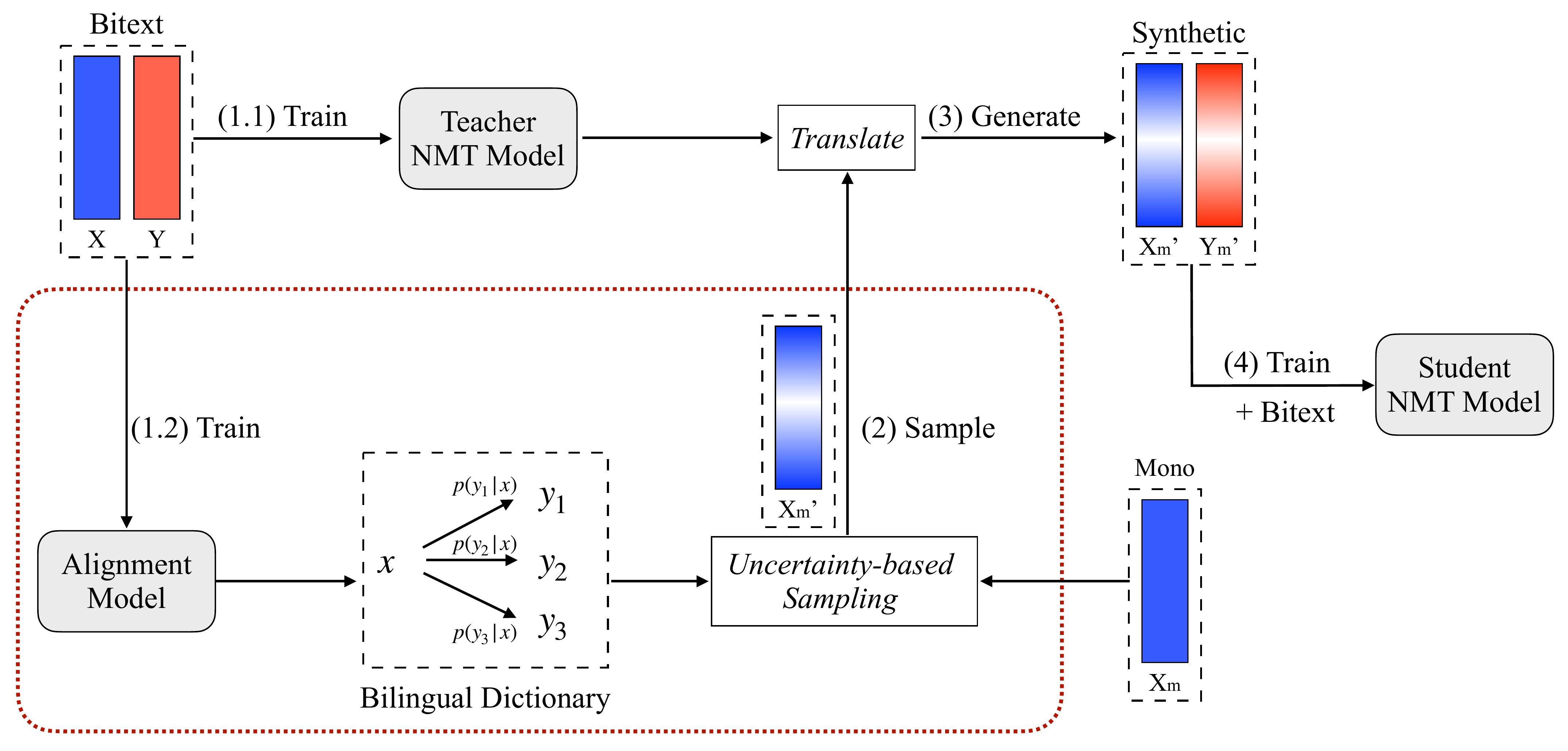}
    \caption{Framework of the proposed uncertainty-based sampling strategy for self-training. Procedures framed in the red dashed box corresponds to our approach integrated into the standard self-training framework. ``Bitext'', ``Mono'', ``Synthetic'' denotes authentic parallel data, monolingual data and synthetic parallel data, respectively. }
    \label{fig:framework}
\end{figure*}

\subsection{Uncertainty-based Sampling Strategy}
\label{sec:sampling_strategy}

With the aforementioned measure of monolingual data uncertainty in Section~\ref{sec:identification_unc}, we propose the uncertainty-based sampling strategy for self-training, which prefers to sample monolingual sentences with relatively high uncertainty. 

To ensure the data diversity and avoid the risk of being dominated by the excessively uncertain sentences, we sample monolingual sentences according to the uncertainty distribution with the highest uncertainty penalized. Specifically, given a budget of $N_s$ sentences to sample, we set two hyper-parameters to control the sampling probability as follows: 
\begin{align}
p &= \frac{\left[\alpha\cdot{\rm U}({\bf x}^{j}|\mathcal{A}_b)\right]^\beta}{\sum_{{\bf x}^{j}\in\mathcal{M}_{x}}\left[\alpha\cdot{\rm U}({\bf x}^{j}|\mathcal{A}_b)\right]^\beta},
\\
\alpha &=
    \begin{cases}
        ~ 1,\quad {\rm U}({\bf x}^{j}|\mathcal{A}_b) \leq {\rm U}_{max}, \\
        max(\frac{2{\rm U}_{max}}{{\rm U}({\bf x}^{j}|\mathcal{A}_b)}-1, 0),\ \ \mathrm{else},
    \end{cases}
\end{align}
where $\alpha$ is used to penalize excessively high uncertainty over a maximum uncertainty threshold ${\rm U}_{max}$ (See Figure~\ref{fig:curves-sampprob}(a)), the power rate $\beta$ is used to adjust the distribution such that a larger $\beta$ gives more probability mass to the sentences with high uncertainty (See Figure~\ref{fig:curves-sampprob}(b)). 

The maximum uncertainty threshold ${\rm U}_{max}$ is assigned to the uncertainty value such that $R\%$ of sentences in the authentic parallel corpus have monolingual data uncertainty below than it. $R$ is assumed to be as high as 80 to 100. Because for monolingual data with uncertainty higher than this threshold, they may not be translated correctly by the ``teacher" model as there are inadequate such sentences in the authentic parallel data for the model to learn.
As a result, monolingual sentences with uncertainty higher than ${\rm U}_{max}$ should be penalized in terms of the sampling probability.

\paragraph{Overall Framework.}
Figure~\ref{fig:framework} presents the framework of our uncertainty-based sampling for self-training, which includes four steps: 1) train a ``teacher" NMT model and an alignment model on the authentic parallel data simultaneously; 2) extract the bilingual dictionary from the alignment model and perform uncertainty-based sampling for monolingual sentences; 3) use the ``teacher" NMT model to translate the sampled monolingual sentences to construct the synthetic parallel data;
4) train a ``student" NMT model on the combination of synthetic and authentic parallel data.

\subsection{Constrained Scenario}
\label{sec:constrained}

We first validated the proposed sampling approach in a constrained scenario, where we followed the experimental configuration in Section~\ref{sec:uncertain_part_performance} with the \textsc{Transformer-Base} model, the 8M bitext, and the 40M monolingual data. It allows the efficient evaluation of our approach with varied combinations of hyper-parameters and also the comparison with related methods. Specifically, we performed our approach by sampling 8M sentences from the 40M monolingual data and then combining the corresponding 8M synthetic data with the 8M bitext to train the \textsc{Transformer-Base} model.

Table~\ref{tab:bleu-hyper} reported the impact of $\beta$ and $R$ on the BLEU score. As shown, sampling with high uncertainty sentences and penalizing those with excessively high uncertainty improves translation performance from 36.6 to 36.9. In these experiments, the uncertainty threshold ${\rm U}_{max}$ for penalizing are 2.90 and 2.74, which are determined by the 90\% and 80\% ($R$=90 and 80 in  Table~\ref{tab:bleu-hyper}) most certain sentences in the authentic parallel data, respectively.  Obviously, the proposed uncertainty-based sampling strategy achieves the best performance with $R$ at 90 and $\beta$ at 2. In the following experiments, we use
$R=90$ and $\beta=2$ as the default setting for our sampling strategy  if not otherwise stated.

\begin{table}[t!]
\fontsize{10}{11}\selectfont
\centering
\begin{tabular}{l c c c c}
    \toprule
     \multicolumn{2}{l}{}\multirow{2}{*}{BLEU}\quad  & \multicolumn{3}{c}{ $R$} \\
     \cmidrule(lr){3-5}
     & & 100 & 90 & 80 \\
     \midrule
    \multirow{3}{*}{$\beta$}  & 1 & 36.6 & 36.7 & 36.6\\
                               & 2 & 36.7 & \bf 36.9 & 36.6  \\
                               & 3 & 36.5 & 36.5 & 36.5  \\
  \bottomrule     
  \end{tabular}
  \caption{Translation performance with respect to different values of $\beta$ and $R$. The BLEU score is averaged on WMT En$\Rightarrow$De newstest2019 and newstest2020.}
  \label{tab:bleu-hyper}
\end{table}

\begin{table*}[t!]
\setcounter{table}{3}
\centering
    \begin{tabular}{l l l l l l l l l}
    \toprule
    \multirow{2}{*}{\bf System} 
    &\multirow{2}{*}{\bf Data} 
    & \multicolumn{3}{c}{\bf  En$\Rightarrow$De} & \multicolumn{3}{c}{\bf  En$\Rightarrow$Zh} \\
    \cmidrule(lr){3-5}\cmidrule(lr){6-8} 
    &  & {\bf 2019} &  {\bf 2020} & {\bf Avg} & {\bf 2019} &  {\bf 2020} & {\bf Avg}\\
    \midrule
    \multirow{2}{*}{\newcite{Wu:2019:exploiting}} & \textsc{Bitext} &  37.3  & -- & -- &-- & -- & -- \\
    & ~~+\textsc{RandSamp}  &  39.8  & -- & -- & -- & -- & -- \\
    \multirow{2}{*}{\newcite{Shi:2020:oppo}} & \textsc{Bitext} & -- & --  & -- & -- & 38.6 & -- \\
    & ~~+\textsc{RandSamp}  & -- & --  & -- & -- & 41.9 & -- \\
    \midrule
    \midrule
    \multirow{3}{*}{\it This Work}
    & \textsc{Bitext}               & 39.6 & 31.0 & 35.3 & 37.1 & 42.5 & 39.8\\
    & ~~+\textsc{RandSamp}         & 41.6 & 33.1 & 37.3 &  37.6 & 43.8 & 40.7\\
    & ~~+\textsc{SrcLM}       & 41.7 & 33.1 & 37.4 & 37.3 & 44.0 & 40.7 \\
    & ~~+\textsc{UncSamp}      &  42.5$^\Uparrow$ & 34.4$^\Uparrow$ & \bf  38.4 & 38.2$^\Uparrow$ & 44.3$^\uparrow$  & \bf 41.3 \\
    \bottomrule
    \end{tabular}
  \caption{Translation performance on WMT En$\Rightarrow$De and WMT En$\Rightarrow$Zh test sets. The results are reported with de-tokenized case-sensitive SacreBLEU. We adopt the \textsc{Transformer-big} with large batch training~\cite{Ott:2018:analyzing} to achieve the strong performance. ``$\uparrow/\Uparrow$'':  indicate statistically significant improvement over \textsc{RandSamp}  $p < 0.05/0.01$ respectively. } 
  \label{tab:main}
\end{table*}

\paragraph{Effect of Sampling.}
Some researchers may doubt that the final translation quality is affected by the quality of the teacher model. Therefore, translations of high-uncertainty sentences should contain many errors, and it is better to add the results of oracle translations to discuss the sampling effect and the quality of pseudo-sentences separately. 
To dispel the doubt, we still used the aforementioned 8M bitext as the bilingual data, and used the rest of WMT19 En-De data (28.8M) as the held-out data (with oracle translations) for sampling. The results are listed in Table~\ref{tab:bleu-oracle}.

Clearly, our uncertainty-based sampling strategy (\textsc{UncSamp}) outperforms the random sampling strategy (\textsc{RandSamp}) when manual translations are used (Rows 2 vs. 3), demonstrating the effectiveness of our sampling strategy based on the uncertainty.
Another interesting finding is that using the pseudo-sentences outperforms using the manual translations (Rows 4 vs. 2, 5 vs. 3). One possible reason is that the \textsc{Transformer-big} model to construct the pseudo-sentences was trained on the whole WMT19 En-De data that contains the held-out data, which serves as self-training to decently improve the supervised baseline~\cite{He:2019:revisiting}.

\begin{table}[t!]
\fontsize{10}{11}\selectfont
\setcounter{table}{1}
\centering
\begin{tabular}{l l c c c}
     \toprule
     \bf\# & \bf Data & \bf 2019 & \bf 2020 & \bf Avg \\
     \midrule
     1 & \textsc{Bitext} & 36.9 & 27.7 & 32.3 \\
     2 & ~+ \textsc{RandSamp Ora} & 37.4 & 28.0 & 32.7 \\
     3 & ~+ \textsc{UncSamp Ora} & 37.8 & 28.2 & \uline{33.0} \\
     4 & ~+ \textsc{RandSamp ST} & 40.0 & 30.1 & 35.0 \\
     5 & ~+ \textsc{UncSamp ST} & 40.4 & 30.5 & \bf{35.4} \\
     \bottomrule
  \end{tabular}
  \caption{Comparison of our \textsc{UncSamp} and \textsc{RandSamp} with manual translations (Ora: manual translations; ST: pseudo-sentences) on WMT En$\Rightarrow$De newstest2019 and newstest2020. }
  \label{tab:bleu-oracle}
\end{table}

\paragraph{Comparison with Related Work.}
We compared our sampling approach with two related works, i.e., difficult word by frequency~\cite[\textsc{DWF},][]{Fadaee:2018:back} and source language model~\cite[\textsc{SrcLM},][]{lewis2010intelligent}. The former one was proposed for monolingual data selection for back-translation, in which sentences with low-frequency words were selected to boost the performance of back-translation. The latter one was proposed for in-domain data selection for in-domain language models. Details of the implementation of related work are in Appendix~\ref{sec:appendix_related}.


Table~\ref{tab:bleu-related} listed the results. 
For DWF, it brings no improvement over \textsc{RandSamp}, indicating that the technique developed for back-translation may not work for self-training.  
As for \textsc{SrcLM}, it achieves a marginal improvement over \textsc{RandSamp}. The proposed \textsc{UncSamp} approach outperforms the baseline \textsc{RandSamp} by +0.7 BLEU point, which demonstrates the effectiveness of our approach. In addition to our \textsc{UncSamp} approach, we also utilized another \textsc{n}-gram language model at the target side to further filter out the synthetic data with potentially erroneous target sentences. By filtering out 20\% sentences from the sampled 8M sentences, our \textsc{UncSamp} approach achieves a further improvement up to +0.9 BLEU point.

\begin{table}[t!]
\fontsize{10}{11}\selectfont
\setcounter{table}{2}
\centering
\begin{tabular}{l c c c}
     \toprule
     \bf Data & \bf 2019 & \bf 2020 & \bf Avg \\
     \midrule
     \textsc{RandSamp}  & 40.9 & 31.6 & 36.2 \\
     \textsc{DWF}  & 39.6 & 30.1 & 34.8 \\
     \textsc{SrcLM}  & 41.1 & 32.0 & 36.5 \\
     \midrule
    \textsc{UncSamp} & 41.6 & 32.3 & \uline{36.9} \\
     ~~~+ Filtering & 41.5 & 32.7 & \bf37.1 \\
     \bottomrule
  \end{tabular}
  \caption{Comparison of the proposed uncertainty-based sampling strategy with related methods on WMT En$\Rightarrow$De newstest2019 and newstest2020.
  }
  \label{tab:bleu-related}
\end{table}

\subsection{Unconstrained Scenario}
\label{sec:unconstrained}
We extended our sampling approach to the unconstrained scenario, where the scale of data and the capacity of NMT models for self-training are increased significantly. We conducted experiments on the high-resource En$\Rightarrow$De and En$\Rightarrow$Zh translation tasks with all the authentic parallel data, including 36.8M sentence pairs for En$\Rightarrow$De and 22.1M for En$\Rightarrow$Zh, respectively. For monolingual data, we considered all the 200M English newscrawl monolingual data to perform sampling. We trained the \textsc{Transformer-Big} model for experiments. 

Table~\ref{tab:main} listed the main results of large-scale self-training on high-resource language pairs. As shown, our \textsc{Transformer-big} models trained on the authentic parallel data achieve the performance competitive with or even better than the submissions to WMT competitions. Based on such strong baselines, self-training with \textsc{RandSamp} improves the performance by +2.0 and +0.9 BLEU points on En$\Rightarrow$De and En$\Rightarrow$Zh tasks respectively, demonstrating the effectiveness of the large-scale self-training for NMT models. With our uncertainty-based sampling strategy \textsc{UncSamp}, self-training achieves further significant improvement by +1.1 and +0.6 BLEU points over the random sampling strategy, which demonstrates the effectiveness of exploiting uncertain monolingual sentences.

\subsection{Analysis}
\label{sec:analysis}

In this section, we conducted analyses to understand how the proposed uncertainty-based sampling approach improved the translation performance. Concretely, we analyzed the translation outputs of WMT En$\Rightarrow$De newstest2019 from the \textsc{Transformer-Big} model in Table~\ref{tab:main}.

\paragraph{Uncertain Sentences.}
As we propose to enhance high uncertainty sentences in self-training, one remaining question is whether our \textsc{UncSamp} approach improves the translation quality of high uncertainty sentences. Specifically, we ranked the source sentences in the newstest2019 by the monolingual uncertainty, and divided them into three equally sized groups, namely Low, Medium and High uncertainty.

The translation performance on these three groups is reported in Table~\ref{tab:analysis-output-uncertainty}. The first observation is that sentences with high uncertainty are with relatively low BLEU scores (i.e., 31.0), indicating the higher difficulty for NMT models to correctly decode the source sentences with higher uncertainty. Our \textsc{UncSamp} approach improves the translation performance on all sentences, especially on the sentences with high uncertainty (+10.9\%), which confirms our motivation of emphasizing the learning on uncertain sentences for self-training.

\begin{table}[t!]
\setcounter{table}{4}
\fontsize{10}{11}\selectfont
\centering
\begin{tabular}{l c c c r}
     \toprule
     \multirow{2}{*}{\bf Unc} 
     & \multirow{2}{*}{\bf \textsc{Bitext}}
     & \multirow{2}{*}{\bf \textsc{RandSamp}}
     &\multicolumn{2}{c}{\bf \textsc{UncSamp}}\\
     \cmidrule(lr){4-5}
     & & &  \bf BLEU & \bf $\triangle(\%)$ \\
     \midrule
     Low & 38.1 & 39.7  & 41.5 & 8.9 \\
     Med & 34.2 & 36.7  & 37.4 & 9.3 \\
     High & 31.0 & 33.4 & 34.4 & \bf10.9 \\
     \bottomrule
\end{tabular}
    \caption{Translation performance on uncertain sentences. The relative improvements over \textsc{Bitext} for \textsc{UncSamp} are also presented.}
    \label{tab:analysis-output-uncertainty}
\end{table}

\paragraph{Low-Frequency Words.} 
Partially motivated by \newcite{Fadaee:2018:back}, we hypothesized that the addition of monolingual data in self-training has the potential to improve the prediction of low-frequency words at the target side for the NMT models. Therefore, we investigated whether our approach has a further boost to the performance on the prediction of low-frequency words.
We calculated the word accuracy of the translation outputs with respect to the reference in newstest2019 by \texttt{compare-mt}. Following~\newcite{Wang:2020:ACL}, we divided words into three categories based on their frequency, including High: the most {3,000} frequent words; Medium: the most {3,001}-{12,000} frequent words; Low: the other words. 

Table~\ref{tab:analysis-output-freq} listed the results of word accuracy on these three groups evaluated by F-measure.
First, we observe that low-frequency words in \textsc{Bitext} are more difficult to predict than medium- and high-frequency words (i.e., 52.3 v.s. 65.2 and 70.3), which is consistent with \newcite{Fadaee:2018:back}. Second, adding monolingual data by self-training improves the prediction performance of low-frequency words. Our \textsc{UncSamp} approach outperforms \textsc{RandSamp} significantly on the low-frequency words. These results suggest that emphasizing the learning on uncertain monolingual sentences also brings additional benefits for the learning of low-frequency words at the target side.

\begin{table}[t!]
\fontsize{10}{11}\selectfont
\centering
\begin{tabular}{l c c c r}
     \toprule
     \multirow{2}{*}{\bf Freq}
     & \multirow{2}{*}{\bf \textsc{Bitext}}
     & \multirow{2}{*}{\bf \textsc{RandSamp}}
     &\multicolumn{2}{c}{\bf \textsc{UncSamp}}\\
     \cmidrule(lr){4-5}
     & & &  \bf Fmeas & \bf $\triangle(\%)$ \\
     \midrule
     Low   & 52.3 & 53.8 & 54.7 & \bf4.5 \\
     Med   & 65.2 & 66.5 & 66.9 & 2.6 \\
     High  & 70.3 & 71.6 & 72.0 & 2.4 \\
     \bottomrule
\end{tabular}
    \caption{Prediction accuracy of low-frequency words in the translation outputs. The relative improvements over \textsc{Bitext} for \textsc{UncSamp} are also presented.}
    \label{tab:analysis-output-freq}
\end{table}

\section{Related Work}
\paragraph{Synthetic Parallel Data.}
Data augmentation by synthetic parallel data has been the most simple and effective way to utilize monolingual data for NMT, which can be achieved by self-training~\cite{He:2019:revisiting} and back-translation~\cite{Sennrich:2016:BT}. While back-translation has dominated the NMT area for years~\cite{Fadaee:2018:back,Edunov:2018:understanding,Caswell:2019:tagged}, recent works on translationese~\cite{Marie:2020:tagged,graham2019translationese} suggest that NMT models trained with back-translation may lead to distortions in automatic and human evaluation. To address the problem, starting from WMT2019~\cite{Barrault:2019:findings}, the test sets only include naturally occurring text at the source-side, which is a more realistic scenario for practical translation usage. In this new testing setup, the forward-translation~\cite{zhang2016exploiting}, i.e., self-training in NMT, becomes a more promising method as it also introduces naturally occurring text at the source-side.
Therefore, we focus on the data sampling strategy in the self-training scenario, which is different from these prior studies.

\paragraph{Data Uncertainty in NMT.} 
Data uncertainty in NMT has been investigated in the last few years. \newcite{Ott:2018:analyzing} analyzed the NMT models with data uncertainty by observing the effectiveness of data uncertainty on the model fitting and beam search. \newcite{Wang:2019:improving} and \newcite{Zhou:2020:uncertainty} computed the data uncertainty on the back-translation data and the authentic parallel data and proposed uncertainty-aware training strategies to improve the model performance, respectively. \newcite{wei2020uncertainty} proposed the uncertainty-aware semantic augmentation method to bridge the discrepancy of the data distribution between the training and the
inference phases. In this work, we propose to explore monolingual data uncertainty to perform data sampling for the self-training in NMT.

\section{Conclusion}

In this work, we demonstrate the necessity of distinguishing monolingual sentences for self-training in NMT, and propose an uncertainty-based sampling strategy to sample monolingual data. By sampling monolingual data with relatively high uncertainty, our method outperforms random sampling significantly on the large-scale WMT English$\Rightarrow$German and English$\Rightarrow$Chinese datasets.
Further analyses demonstrate that our uncertainty-based sampling approach does improve the translation quality of high uncertainty sentences and also benefits the prediction of low-frequency words at the target side. 
The proposed technology has been applied to TranSmart\footnote{\url{https://transmart.qq.com/index}}~\cite{transmart2021}, an interactive machine translation system in Tencent, to improve the performance of its core translation engine.
Future work includes the investigation on the confirmation bias issue of self-training and the effect of decoding strategies on self-training sampling.

\section*{Acknowledgments}

This work is partially supported by the Research Grants Council of the Hong Kong Special Administrative Region, China (CUHK 2410021, Research Impact Fund (RIF), R5034-18; CUHK 14210717, General Research Fund), and Tencent AI Lab RhinoBird Focused
Research Program (GF202036). We sincerely thank the anonymous reviewers for their insightful suggestions on various aspects of this work.

\bibliographystyle{acl_natbib}
\bibliography{anthology,acl2021}

\appendix
\section{Appendix}
\label{sec:appendix}

\subsection{Synthetic Data}
\label{sec:appendix_syn}

When performing self-training, we constructed the synthetic data by translating the monolingual sentences via beam search with beam width $5$, and followed~\newcite{Edunov:2018:understanding}\footnote{\url{https://github.com/pytorch/fairseq/tree/master/examples/backtranslation}} to remove sentences longer than $250$ words as well as sentence-pairs with a source/target length ratio exceeding $1.5$. The ``teacher" NMT model for self-training is the \textsc{Transformer-Big} model to ensure the quality of synthetic data.

\subsection{Linguistic Properties}
\label{sec:appendix_ling}

\paragraph{Word Rarity.}
Word rarity measures the frequency of words in a sentence with a higher value indicating a more rare sentence~\cite{Platanios:2019:competence}. The word rarity of a sentence is calculated as follows:
\begin{align}
    {\rm WR}({\bf x}) = - \frac{1}{T_x}\sum_{t=1}^{T_x}\log{p(x_t)},
\end{align}
where $p(x_t)$ denotes the normalized frequency of word $x_t$ in the authentic parallel data, and $T_x$ is the sentence length. 

\paragraph{Coverage.}
Coverage measures the ratio of source words being aligned by any target words~\cite{tu2016modeling}. Firstly, we trained an alignment model on the authentic parallel data by \emph{fast-align}\footnote{\url{https://github.com/clab/fast\_align}}. Then we used the alignment model to force-align the monolingual sentences and the synthetic target sentences. Next, we calculated the coverage of each source sentence, and report the averaged coverage of each data bin. The lower coverage of monolingual sentences in bin 5 indicates that they are not aligned as well as the other bins.

\subsection{Comparison with Related Work}
\label{sec:appendix_related}

We compared our sampling approach with two related works, i.e., difficult word by frequency~\cite[\textsc{DWF},][]{Fadaee:2018:back} and source language model~\cite[\textsc{SrcLM},][]{lewis2010intelligent}. The former one was proposed for monolingual data selection for back-translation, in which sentences with low-frequency words were selected to boost the performance of back-translation. The latter one was proposed for in-domain data selection for in-domain language models.

For \textsc{DWF}, we ranked the monolingual data by word rarity~\cite{Platanios:2019:competence} of sentences and also selected the top 80M monolingual data for self-training. For \textsc{SrcLM}, we trained an \textsc{n}-gram language model~\cite{heafield2011kenlm}\footnote{\url{https://kheafield.com/code/kenlm/}} on the source sentences in the bitext and measured the distance between each monolingual sentence to the bitext source sentences by cross-entropy. Similarly, we selected 8M monolingual data with the lowest cross-entropy for self-training.

\end{document}